\documentclass[letterpaper, 10 pt, conference]{ieeeconf}

\IEEEoverridecommandlockouts
\usepackage{amsmath,amsfonts}
\usepackage{textcomp}
\usepackage{stfloats}
\usepackage{graphicx} 
\usepackage{bm}
\usepackage{subfig}
\usepackage{xcolor}
\usepackage{array}
\usepackage{url}
\usepackage{hyperref}
\newcolumntype{P}[1]{>{\centering\arraybackslash}p{#1}}
\newcolumntype{M}[1]{>{\centering\arraybackslash}m{#1}}
\usepackage{booktabs}
\usepackage{xcolor,colortbl}
\definecolor{lavender}{rgb}{0.9, 0.9, 0.98}

\title{\LARGE \bf

Unidirectional Human-Robot-Human Physical Interaction \\for Gait Training
}

\author{Lorenzo Amato$^{1,2}$, Lorenzo Vianello$^{2}$, Emek Barış Küçüktabak$^{2,3}$, Clément Lhoste$^{2}$, Matthew Short$^{2,4}$, \\ Daniel Ludvig$^{4}$,  Kevin Lynch$^{3}$, Levi Hargrove$^{5,6}$, Jose L. Pons$^{2,3,4,6}$%
\thanks{$^1$ The Biorobotics Institute, Scuola Superiore Sant’Anna, 56025 Pontedera, Italy \textit{and} Department of Excellence in Robotics \& AI, Scuola Superiore Sant'Anna, 56127 Pisa, Italy}
\thanks{$^2$ Legs and Walking Lab of Shirley Ryan AbilityLab, Chicago, IL, USA}
\thanks{$^3$ Center for Robotics and Biosystems of Northwestern University, Evanston, IL, USA}
\thanks{$^4$ Department of Biomedical Engineering, Northwestern University, Evanston, IL, USA}
\thanks{$^5$ Neural Engineering for Prosthetics and Orthotics Lab of Shirley Ryan AbilityLab, Chicago, IL, USA}
\thanks{$^6$ Department of Physical Medicine and Rehabilitation, Northwestern University, Chicago, IL, USA}
}

\begin{document}
\maketitle

\begin{abstract}

This work presents a novel rehabilitation framework designed for a therapist, wearing an inertial measurement unit (IMU) suit, to virtually interact with a lower-limb exoskeleton worn by a patient with motor impairments. This framework aims to harmonize the skills and knowledge of the therapist with the capabilities of the exoskeleton. The therapist can guide the patient's movements by moving their own joints and making real-time adjustments to meet the patient's needs, while reducing the physical effort of the therapist. This eliminates the need for a predefined trajectory for the patient to follow, as in conventional robotic gait training.
For the virtual interaction medium between the therapist and patient, we propose an impedance profile that is stiff at low frequencies and less stiff at high frequencies, that can be tailored to individual patient needs and different stages of rehabilitation.
The desired interaction torque from this medium is commanded to a whole-exoskeleton closed-loop compensation controller. The proposed virtual interaction framework was evaluated with a pair of unimpaired individuals in different teacher-student gait training exercises. Results show the proposed interaction control effectively transmits haptic cues, informing future applications in rehabilitation scenarios. 

\end{abstract}

\section{Introduction}

Gait recovery is a major objective in the rehabilitation of patients with lower-limb impairments.
Current interventions are largely based on repetitive gait training administered by physical therapists \cite{belda2011rehabilitation}. Typically, the therapist supports and corrects their patient’s movement patterns via physical assistance.
These approaches require significant physical effort from the therapist which can, over time, lead to work-related injuries \cite{mccrory2014work}. Moreover, there are difficulties in performing multi-joint assessments and mobilizations because the therapist-patient physical interaction is limited to a few contact points; therefore an an objective and continuous assessment of the patient's performance is limited. 

\begin{figure}[t]
    \centering
    \includegraphics[width=1\linewidth]{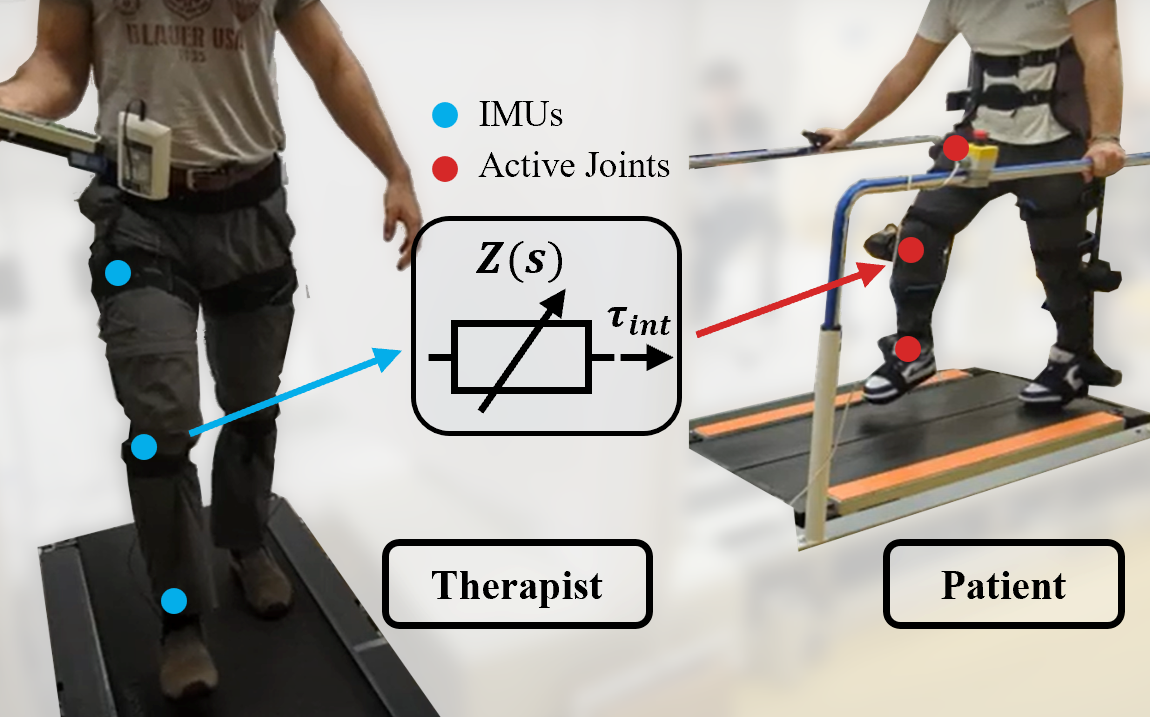}
    \caption{\small \textbf{Framework structure:} A therapist, on the left, can virtually interact with the patient, on the right, through unidirectional impedance interaction control while performing a rehabilitation exercise. The therapist wears the IMU suit, while the patient wears the Exo-H3 exoskeleton.}
    \label{fig:Exo and IMU suit}
    \vspace{-0.5cm}
\end{figure}

Exoskeletons offer a solution to these problems by increasing the intensity of therapy and providing quantifiable measures of patient performance~\cite{belda2011rehabilitation, moeller2023use}. At the same time, they reduce the amount of physical effort required from the therapist, as they can provide joint mobilization and weight support~\cite{gassert2018rehabilitation}. 
To target motor learning, the most common exoskeleton control strategies for gait restoration use either assist-as-needed \cite{belda2011rehabilitation}, especially in early stages of rehabilitation, or error augmentation \cite{marchal2019haptic} approaches, for less impaired patients. These encourage the patient's active motion, adapting robotic assistance based on the patient's performance of different gait patterns \cite{hobbs2020review,baud2021review,marchal2009review}. 
The goal of assist-as-needed strategies is to encourage the user to move toward a predefined trajectory by controlling the mechanical impedance between the desired trajectory and the user's actual movement. However, the definition of appropriate desired limb trajectories is an open question~\cite{belda2011rehabilitation}. 

Common rehabilitation strategies are based on predefined reference gaits, and they typically segment a gait into recurring phases (e.g., stance and swing) and use different control policies during each phase~\cite{baud2021review}.
However, during training for different mobility scenarios including obstacle avoidance, these approaches fail to adapt in real-time. 
Furthermore, they do not leverage the therapist’s knowledge and ability, and they do not adapt to the patient's specific needs, resulting in reduced efficacy of the rehabilitation due to limited therapist agency \cite{celian2021day,hasson2023neurorehabilitation}.

To incorporate the physical therapist's expertise and leverage their adaptability to patients' needs, previous works have proposed the use of physical human-robot-human interaction (pHRHI) in rehabilitation settings \cite{Kucuktabak2021}. In pHRHI, humans are virtually connected to each other through robot(s), often by virtual passive elements such as a spring and a damper \cite{ganesh2014two, short2023haptic}.
The robots become a medium for physical interaction between two or more users. The interaction can be customized, enabling different rehabilitation paradigms (e.g., collaborative, cooperative, competitive) and harmonizing the human's cognitive and motor skills with the robot's capabilities \cite{Kucuktabak2021}. 
For these reasons, previous works have presented virtual haptic rendering between two exoskeletons as a valid tool to mediate interactions between two individuals \cite{kuccuktabak2023virtual, rahman2011tele, sakurai2009development, vianello2024exoskeleton}. In particular, Küçüktabak et al.~\cite{kuccuktabak2023virtual} introduced interactions between two lower-limb exoskeletons, both unidirectional and bidirectional.
However, the use of two exoskeletons presents disadvantages such as increased cost and increased time for donning and doffing equipment.
Solutions have been proposed to overcome these issues~\cite{pavon2020iot,duan2018motion}, even if their application in a rehabilitation environment still remains an open question.
\begin{table*}[t]\centering
\vspace{0.2cm}
\caption{\small
Advantages and disadvantages of different rehabilitation practices for gait training. Each practice is evaluated across different categories in the context of practical implementation into clinical practice. Categories are scored with respect to conventional gait therapy (i.e., overground or treadmill walking with manual assistance from a physical therapist); + and - indicate better and worse scores than conventional (0), respectively.}
\begin{tabular}{ |M{3cm}||M{1.5cm}|M{1.6cm}|M{1.6cm}|M{1.6cm}|M{1.6cm}|M{1.5cm}|M{1.6cm}|  }
 \hline
 Rehabilitation Practice & Therapist's effort (+: less effort) & Leverages therapist's expertise &Preparation time (+: less time) & Performance measurability & Adaptability to different exercises & Multi-joint mobilization & Force feedback to therapist\\
 \hline \\ [-2ex]
 Conventional gait therapy \cite{belda2011rehabilitation}   & 0 & 0 & 0 & 0 & 0 & 0 & 0 \\ 
 Exoskeleton therapy \cite{baud2021review}   & +++ & - & - & + & - & + & -\\
 Bidirectional Exo-mediated interaction \cite{kuccuktabak2023virtual, vianello2024exoskeleton}   & + & 0 & -{}- & +  & 0 & + & +\\
 \rowcolor{lavender}
\textbf{Unidirectional Exo-mediated interaction} [this paper]   & ++ & 0 & - & + & 0 & + & -\\
 \hline
\end{tabular} 
\label{tab:rehabilitation comparison}
\vspace{-0.45cm}
\end{table*}

In this work, we aim to address some of the challenges that may arise in the interaction between two users, with a focus on functional exercises for rehabilitation~\cite{koh2021exploiting}.
We propose a pHRHI framework for gait rehabilitation where conventional therapy and robotic rehabilitation are merged together. In particular, a therapist wearing an IMU suit (left in Fig.~\ref{fig:Exo and IMU suit}) can guide both of the patient's lower limbs simultaneously (right in Fig.~\ref{fig:Exo and IMU suit}), through unidirectional interaction between the IMU suit and the patient's exoskeleton. 
This approach allows therapists to modify their own leg movements to apply corrections to the patient's joint angles without predefined exoskeleton trajectories, facilitating intuitive adaptation to each patient's impairment and fatigue. This eliminates the need for a finite state machine or gait classification algorithm when switching between exercises and can handle unstructured or unexpected scenarios.
Table~\ref{tab:rehabilitation comparison} displays the advantages and disadvantages of this new formulation compared to previous approaches. 

The proposed framework mediates the interaction between the therapist and patient through a virtual impedance that
offers more support and guidance for slow, deliberate movements that are typical of a patient performing rehabilitative exercises. At the same time, this impedance profile smooths quick and abrupt movements which can be uncomfortable and even dangerous for the patient.
Other works have implemented methods to adapt impedance by learning gain values based on the rehabilitation exercise performed \cite{tran2016evaluation, huo2021impedance}. However, to the best of our knowledge, no previous work has implemented an interaction medium between two humans to target specific movement frequencies for rehabilitation.
Finally, we implemented a whole-exoskeleton closed-loop compensation (WECC) controller similar to \cite{baris2024haptic}, adapted for a higher-DoF system, to follow desired interaction torques.

The framework was validated by a pair of healthy users. 
We tested different scenarios where the two users executed a specific walking pattern or avoided obstacles. 
These experiments demonstrate the approach's ability to provide haptic cues at multiple contact points, enabling controlled variability in non-repetitive rehabilitation exercises to promote motor learning. 
The video of the experiments is available \href{https://youtu.be/rKgTC0fmx1Y}{here}.


\section{Methods}

\subsection{Exo-H3 Exoskeleton} \label{Exo-H3 Exoskeleton}
The Exo-H3 exoskeleton (Technaid S.L., Spain) was used as a platform to validate the proposed framework (Fig.~\ref{fig:Exo and IMU suit}, right side). This exoskeleton has six active DoF to mobilize the hip, knee, and ankle joints on the left and right legs in the sagittal plane. Strain gauges embedded at each joint, with 0.1~Nm joint torque resolution, allow for the estimation of the interaction torque with the user. In each footplate, two pressure sensors are equipped to detect contact with the ground.
A computer (Intel Core$^{TM}$ i7-6700H with 4 cores at 3.4GHz, 32GB of RAM) runs the high-level control algorithm at 100~Hz, and it manages the communication with the joint motors and sensors of the robot over two CAN buses. The same computer also manages the communication with the IMU sensors through a ROS network. The individual motor drivers run the low-level controllers at 1~kHz. 

\subsection{IMU suit} \label{IMU suit}
The IMU suit is composed of six IMUs (Tech-IMU V3, Technaid S.L., Spain), one at each thigh, shank, and foot (Fig.~\ref{fig:Exo and IMU suit}, left side). The IMUs are secured to the user's body through straps, allowing fast donning and freedom of movement to the user. The signals are collected by a hub (Tech-HUB V3, Technaid S.L.) and transmitted by a ROS network at 200 Hz to the exoskeleton controller.
The joint angles are calculated as the difference between the sagittal plane angles of consecutive links and used as reference angles ($\bm{\theta_{\text{des}}}$) during IMU-exoskeleton interaction. For the hip angle calculation, we assumed the trunk remained vertical, avoiding an additional IMU on the trunk for simplicity.

\begin{figure*}[t]
    \centering 
    \vspace{0.2cm}
    \includegraphics[width=0.95\linewidth]{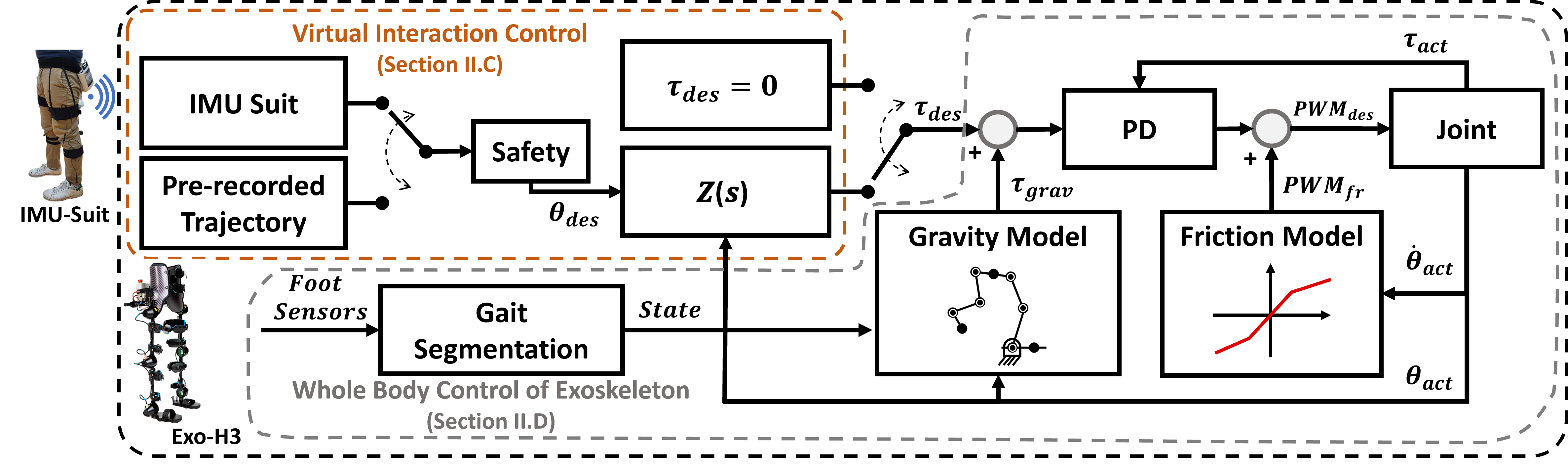}
    \caption{\small Block diagram of the Exo-H3 controller. On the top-left portion, the interaction control computes the desired torque ($\tau_{\text{des}}$) using the impedance control transfer function $Z(s)$. The reference trajectory is defined by the partner selector between pre-recorded trajectories and the IMU-suit angles. In case the interaction control is not selected, it is possible to switch the control modality to transparent control ($\tau_{\text{des}}=0$). On the bottom-right portion, the WECC is responsible for compensating for the exoskeleton's gravity depending on the detected gait state. Then, it compensates for friction and implements the torque controller to drive the robotic joint.}
    \label{fig:Control Architecture}
    \vspace{-0.4cm}
\end{figure*}

\subsection{Virtual Interaction Control} \label{Interaction Control}

Throughout this paper, we refer to the pair of subjects in our experiments as the teacher and the student, wearing the IMU suit and exoskeleton, respectively.
To establish the teacher-student interaction, a virtual interaction control was implemented on the exoskeleton controller. Figure~\ref{fig:Control Architecture} presents the block scheme of the control architecture. The controller defines a desired interaction torque ($ \boldsymbol{\tau}_{\text{des}}$) between the exoskeleton and the student to compliantly follow the teacher's movements.
In particular, an impedance controller computes the desired torque value based on the error between a desired angle ($\theta_{\text{des}}$) and the actual joint angle ($\theta_{\text{act}}$) at each joint. The closer the actual joint angle to the desired, the less assistance is provided by the exoskeleton.

The desired joint angles are defined according to the rehabilitation exercise chosen by a trajectory selector that enables switching between the classic predefined trajectory approach to guide the joint movements and a real-time virtual connection with the IMU suit.

Finally, a safety module adapts the desired joint angle signals to the robot's and patient's range of motion (ROM). The Exo-H3 allows 30~deg of plantar- and dorsiflexion, 105~deg of knee flexion, 5~deg of knee extension, 105~deg of hip flexion, and 30 deg of hip extension. Furthermore, the desired interaction torque rate of change was bounded to 200~Nm/s to avoid abrupt changes in the desired torque interaction values, which are potentially harmful to the patient.

\begin{figure}[t]
    \centering 
    \includegraphics[width=1\linewidth]{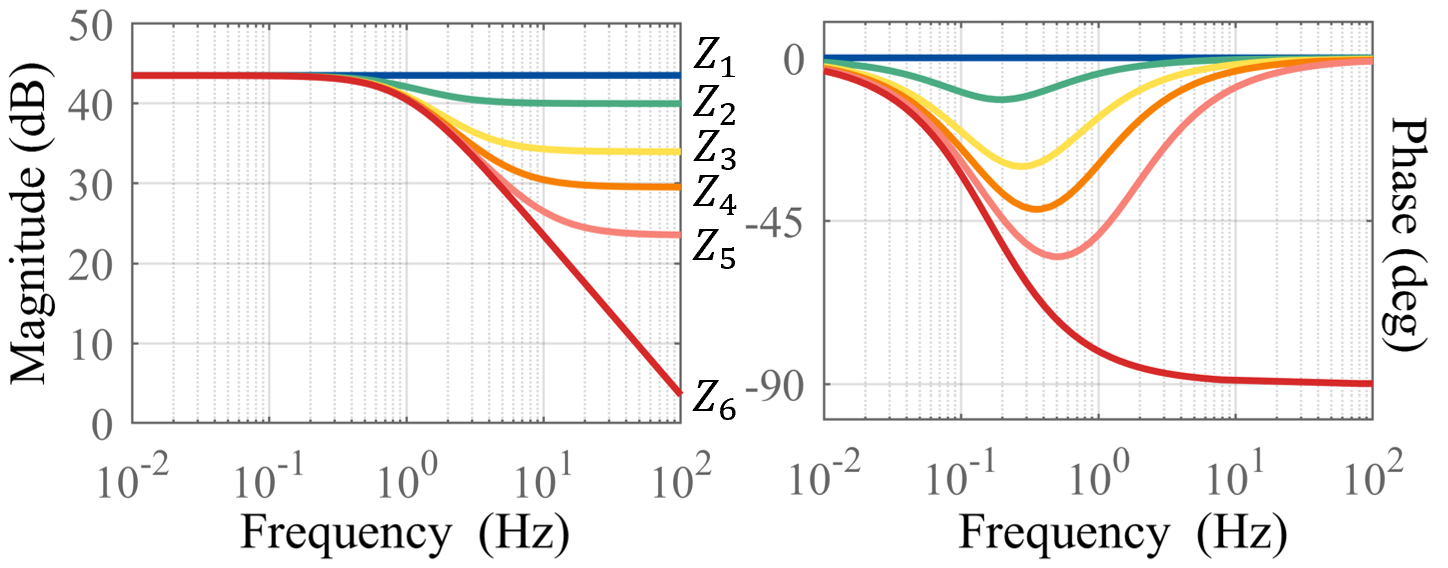}
    \caption{\small Bode plot of different $Z(s)$ changing the $k_{\text{high\_f}}$ values ([$Z_1$, $Z_2$, $Z_3$, $Z_4$, $Z_5$, $Z_6$]=[150, 100, 50, 30, 15, 0]~Nm/rad), while $k_{\text{low\_f}} = 150$~Nm/rad and $f_{\text{cut}}=1$~Hz are kept constant. $Z_1$ represents the constant stiffness case, while $Z_6$ represents the low-pass filter case.}
    \label{fig:bode plot}
    \vspace{-0.5cm}
\end{figure}

In this work, the impedance controller is designed in the frequency domain as a virtual impedance ($Z(s)$) 
and can be expressed for a single joint as:
\begin{equation}
    {\tau_{\text{des}}}(s)={Z}(s) \theta_{\text{e}}(s), \; \text{where} \; \theta_{\text{e}}(s) = {\theta_{\text{des}}}(s)-{\theta_{\text{act}}}(s).
\end{equation}
We choose the impedance profile to be a one-pole, one-zero transfer function, which corresponds to a stiffness at low frequencies ($ k_{\text{low\_f}} $), a stiffness at high frequencies ($ k_{\text{high\_f}} $), and a cut-off frequency ($ f_{\text{cut}} $) at the transition between the two stiffnesses,  
\begin{equation}
\begin{split}
    Z(s) = k_{\text{high\_f}} \frac{(s+f_{\text{cut}}\frac{k_{\text{low\_f}}}{k_{\text{high\_f}}})}{(s+f_{\text{cut}})},
\end{split}
\label{eq:transfer function}
\end{equation}
as illustrated in Fig.~\ref{fig:bode plot}.

The frequency response can be modified by changing the values of $k_{\text{low\_f}}, k_{\text{high\_f}}$, and $f_{\text{cut}}$.
This proposed formulation allows simultaneously having a stiffer connection for low-frequency components of $\theta_e$ and a more compliant connection for the high-frequency components. This interaction impedance can be customized to align with the user's speed and abilities. 
For instance, a stiffer connection may be important in early rehabilitation stages when greater assistance is required.
However, stiff interactions may result in instability when high-frequency components of the tracking error arise. 
In human-human coupling, these high-frequency components may occur due to natural human oscillations or pathological behaviors.
A more compliant connection at high frequencies reduces the chance of instability, improving the overall safety of the coupled system.

These parameters ($k_{\text{high\_f}}, k_{\text{low\_f}}, f_{\text{cut}}$) can be dynamically adjusted during rehabilitation by the therapist according to the level of impairment of the patient.
In particular, it is possible to limit the interaction to have frequency components that are useful for the rehabilitation program, shifting from low frequencies ($k_{\text{high\_f}} > k_{\text{low\_f}}$, $f_{\text{cut}} \approx 1$~Hz) in the early stages of rehabilitation to high frequencies ($k_{\text{high\_f}} < k_{\text{low\_f}}$, $f_{\text{cut}} > 1$ Hz) when the patient is recovering motor functionality.
For instance, when $k_{\text{high\_f}} = 0$~Nm/rad, the proposed approach results in a low-pass filter applied to the controller's input signal ($Z_6$ in Fig.~\ref{fig:bode plot}). Otherwise, by setting $k_{\text{high\_f}} = k_{\text{low\_f}} $, the behavior is a constant stiffness across the whole frequency spectrum ($Z_1$ in Fig.~\ref{fig:bode plot}).

A transparent control ($ \boldsymbol{\tau}_{\text{des}} = 0 $) was also implemented to allow the user to move freely. In this work, this form of interaction was studied to evaluate the performance of the whole-body control when it seeks to minimize interaction with the user and support the weight of the exoskeleton. Additionally, it provides us with a baseline of the user's movement execution, which is used in the evaluation of the patient's performance.

\subsection{Whole Body Control of Exoskeleton} \label{whole body control}
After the desired interaction torque ($ \boldsymbol{\tau}_{\text{des}}$) is defined by the virtual interaction controller as explained in Section~\ref{Interaction Control}, the WECC is used to track this torque.

In previous work, we proposed the WECC to control and estimate user-exoskeleton interaction torques for a floating-base hip-knee exoskeleton with feet \cite{baris2024haptic}. In this work, we extend that approach to an exoskeleton that includes active ankle joints as well. The modified WECC has a whole-body gravity compensation algorithm that calculates exoskeleton joint torques to compensate gravity acting on the exoskeleton. These torques are added to the interaction joint torques calculated by the virtual interaction controller, and the total is sent to a torque PD controller that compares to the actual joint torques and generates pulse-width modulation (PWM) duty cycle percentages for the joint drivers. These percentages are summed with friction compensation terms described below. 

For the gravity compensation algorithm, the exoskeleton is modeled as a floating base, seven-link mechanism with six active joints (Fig.~\ref{fig:Control Architecture}, Gravity Model). In addition to the masses of backpack and feet, the thigh and shank center of masses are assumed to be located at the hip and knee joints, respectively, due to the much higher weight of the Exo-H3 actuators with respect to their connecting links.
Exoskeleton link lengths were measured after fitting the exoskeleton to the user's legs. Following the WECC control algorithm in~\cite{baris2024haptic}, joint torques to compensate gravity acting on the exoskeleton are calculated based in part on the stance state as determined by pressure sensors on the exoskeleton's feet. 


To identify the friction components, a system identification test was performed keeping the links in the horizontal plane to avoid gravitational effects, and commanding to the impedance control a desired position chirp signal from 0.01 Hz up to 3 Hz. To be consistent with the voltage controller on the Exo-H3 joint drivers, the PWM duty cycle percentage values were recorded. These values, with respect to the joint speed, were fitted to a linear friction model, assuming the relation between friction torque and voltage is linear at low speeds as in rehabilitation applications.
The friction compensation was implemented in the low-level controller as a direct input to the motor driver, meaning that it is controller-independent, and minimizes the nonlinear effects of friction.

\subsection{Experimental Protocols}

This section describes the tests to validate unidirectional robot-mediated physical interaction. The results are given in Section~\ref{results}.

\subsubsection{Validation of Transparency Control} \label{Transparent}
Interaction torque tracking performance of the WECC control is first verified as it is a crucial component to accurately render different interactions.

To verify the WECC performance, an unimpaired subject was asked to wear the exoskeleton and to walk in transparent mode for 3 minutes on a treadmill at 0.5 km/h, 1.1 km/h, and 1.5 km/h, similarly to \cite{baris2024haptic}. These relatively slow speeds represent typical walking training speeds in physical rehabilitation settings \cite{louie2015gait, perry1995classification}. The transparency evaluation involved analyzing interaction torques and joint angles: low interaction torque \cite{baris2024haptic} and similar joint kinematics to free walking~\cite{mentiplay2018lower} indicate good transparency.
This preliminary experiment provides the baseline of the user's movement used in the evaluation process. 

\subsubsection{Validation of Interaction Impedance Control} \label{Playback}
The effects of the implemented impedance controller were then verified while enabling interaction with a virtual teacher while the student walked on a treadmill at the speeds of 0.5 km/h, 1.1 km/h, and 1.5 km/h. In each case, the trajectory provided by the ``virtual teacher'' was the average of the trajectories recorded by the student while walking in transparent mode (Section~\ref{Transparent}). At each speed, the student walked with two different impedances, for 3 minutes each, at the exoskeleton's joints: constant stiffness across frequencies ($Z_{\text{constant}}$: $k_{\text{low\_f}} = k_{\text{high\_f}} = 100 $ Nm/rad) and an impedance designed to have reduced stiffness at higher frequencies ($Z_{\text{designed}}$: $k_{\text{low\_f}} = 100 $ Nm/rad, $k_{\text{high\_f}}$ = 50, $f_{\text{cut}} = 1$ Hz). No visual feedback of the desired trajectory was given and the student was asked to minimize interaction forces with the exoskeleton. This virtual teacher represents a common rehabilitation scenario where predefined trajectories are used to guide the patient's movement. The interaction torques and the tracking errors were recorded and used to quantify the performance of the impedance controller.
 
\subsubsection{IMU-Exoskeleton Human-Robot-Human Interaction}
After validating the proposed interaction control, we demonstrated its capabilities during pHRHI in three distinct teacher-student rehabilitation scenarios.
In all three scenarios two unimpaired subjects, one wearing the IMU suit (teacher) 
and one wearing the H3-Exo (student), 
were asked to walk on two different treadmills without seeing each other.

First, we tested the effect of virtual connection stiffness during teacher-student interaction during natural walking.
The two subjects were asked to walk at 1.1 km/h, while three different interaction impedances were applied to the exoskeleton controller: a) no interaction ($Z_{\text{zero}}$: $k_{\text{low\_f}} = k_{\text{high\_f}}=0$, $f_{\text{cut}}=1$), 
b) soft interaction ($Z_{\text{soft}}$: $k_{\text{low\_f}}=100$, $k_{\text{high\_f}}=50$, $f_{\text{cut}}=1$), 
and c) stiff interaction ($Z_{\text{stiff}}$: $k_{\text{low\_f}}=200$, $k_{\text{high\_f}}=100$, $f_{\text{cut}}=1$). Based on previous work involving single-joint haptic interaction with limited visual feedback~\cite{short2023haptic}, we hypothesize that stiffer connections will more closely align the joint configurations of the student and the teacher. For this reason, the correlation between the joint kinematics of the two users was evaluated in all conditions.
 
In the second scenario, the proposed approach was used to evaluate the effect of a stiff connection ($Z_{\text{stiff}}$) in teaching an asymmetric walking pattern. 
This teacher-student scenario can be useful especially in the early stages of rehabilitation where a natural gait pattern is too challenging for the patient. The therapist can adapt the gait pattern depending on the specific patient's needs, providing online haptic corrections to the patient while walking. Moreover, our exoskeleton-mediated interaction approach allows for corrections to multiple joints simultaneously by leveraging the synergies between the links involved in the movement.
The teacher was asked to perform a consistent asymmetric gait, where the movement of the left leg had a larger ROM relative to the right. The student's exoskeleton was controlled in interaction control with stiff interaction; he was asked to change his pattern to reproduce the teacher's movement relying only on the haptic feedback. The teacher and student walked together at 0.7 km/h for three minutes.
Kinematics and tracking performance of both subjects were evaluated. 

In the third scenario, pHRHI was evaluated during an obstacle avoidance task while walking at 0.7 km/h. 
The teacher was asked to step over randomly placed physical obstacles, while the student's exoskeleton was controlled in interaction control with stiff interaction ($Z_{\text{stiff}}$) and he was asked to adapt his movement to follow the suggested movements.
These physical obstacles were perceived by the student as a random variation of the natural gait pattern. For this reason, foot clearance was compared between the users to investigate if the proposed control can promptly respond and induce changes in the gait pattern when random events are encountered. In particular, the proposed scenario can represent random disturbances or unstructured rehabilitation exercises that are difficult to facilitate while using exoskeletons with predefined trajectories, and where a prompt response from the therapist is required.
Introducing obstacles to the therapist's gait induces a movement in the patient that we refer to as 'virtual obstacles.' This can promote movement variability (essential for motor learning) without having to interact with physical obstacles that could pose a risk to the patient.

\subsection{Data Recording and Analysis}
All exoskeleton and IMU data were collected at 100 Hz. During offline processing, the datasets were segmented using the heel strikes of the corresponding leg. Each segmented gait step, for each condition, was then expressed as percentage of gait cycle to easily compare and average with the other steps. 
The correlation coefficients between teacher and student joint trajectories were computed using the \texttt{corr} function in MATLAB.

\section{Results and Discussion } \label{results}

\subsection{Validation of Transparency Control}

\begin{figure}[t]
    \centering 
    \includegraphics[width=1\linewidth]{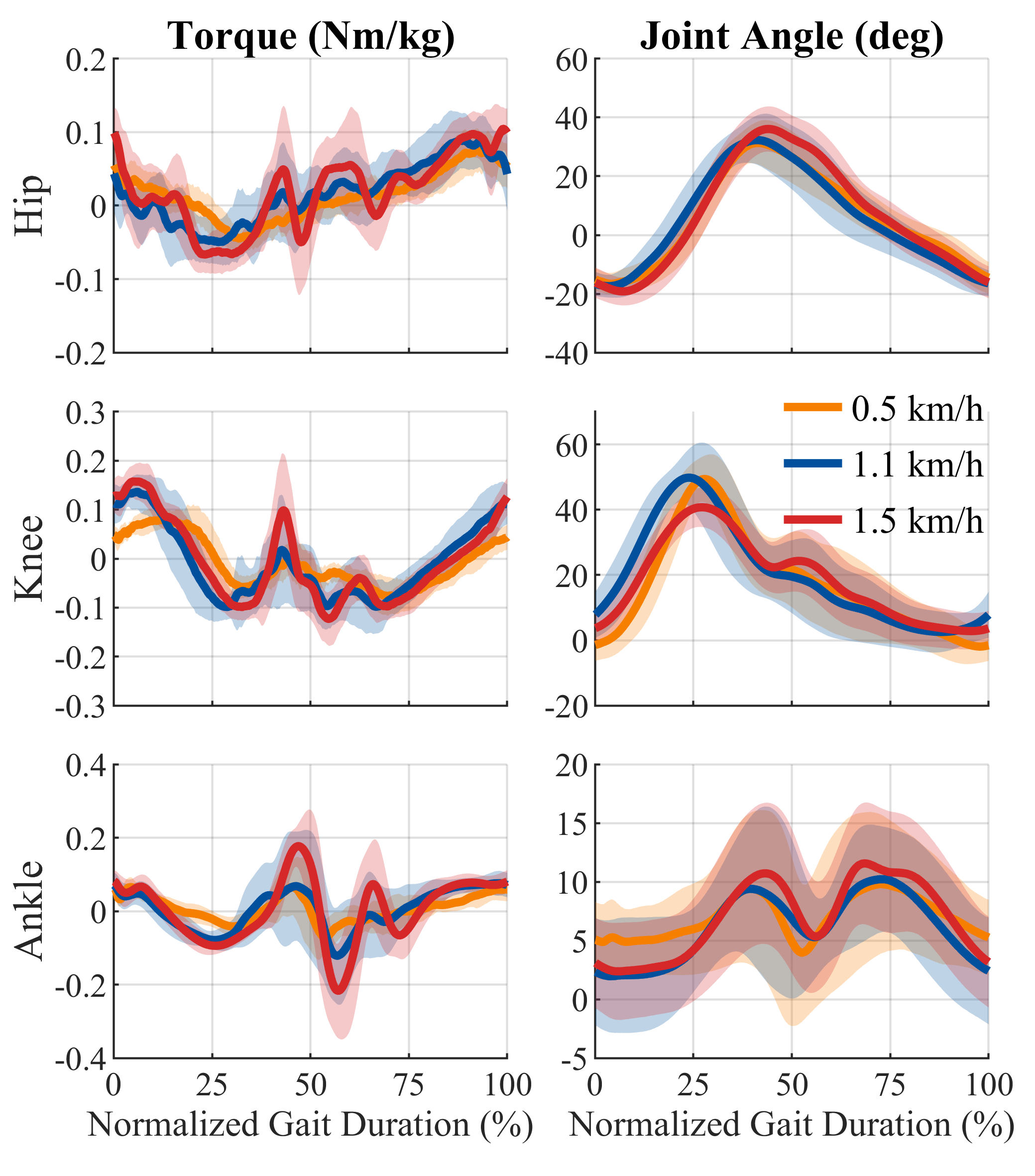}
    \caption{\small Transparency results of the WECC while walking at 0.5 km/h, 1.1 km/h, and 1.5 km/h. The left column shows the mean interaction torque normalized by the subject's body mass, while the right column shows the average joint angles. The shaded areas represent +/- one standard deviation.}
    \label{fig:Results transparency}
    \vspace{-0.5cm}
\end{figure}

In this section, we discuss the performance of the WECC. The results show low interaction torque errors, comparable to our previous work \cite{baris2024haptic}.

Figure~\ref{fig:Results transparency} shows the mean interaction torque (left column) normalized by the subject body mass at different speeds during the gait cycle.
For all the speeds, the WECC showed consistent average transparency performance during the gait cycle, with interaction torques mostly bounded to the range of $\pm$0.15 Nm/kg, and with an increasing trend for faster speeds. 
In particular, for the 0.5 km/h speed, the mean absolute interaction torque was 0.03$\pm$0.020 Nm/kg, 0.05$\pm$0.022 Nm/kg, and 0.03$\pm$0.019 Nm/kg at the hip, knee, and ankle joints, respectively. 
For the 1.1 km/h speed, it was 0.04$\pm$0.025 Nm/kg, 0.06$\pm$0.035 Nm/kg, and 0.05$\pm$0.026 Nm/kg, while for the 1.5 km/h speed it was 0.04$\pm$0.028 Nm/kg, 0.07$\pm$0.040 Nm/kg, and 0.07$\pm$0.047 Nm/kg.
Joint angles with respect to the normalized gait duration are presented in the right column of Fig.~\ref{fig:Results transparency}. Each joint showed similar ROM and shape with respect to healthy individuals' gait angles \cite{mentiplay2018lower}. These angles were then used in the playback test as reference trajectories at the corresponding speed (Section \ref{Playback}).
In general, the WECC resulted in comparable transparency performance to the previous work of \cite{baris2024haptic}, which demonstrated the benefits of the WECC approach compared to common approaches used in literature. 
Moreover, it is possible to conclude that the implemented WECC can properly follow the commanded desired interaction torque, allowing the desired haptic behavior to be properly rendered to the exoskeleton user.

\subsection{Validation of Interaction Impedance Control}
In this section, we present the results of the evaluation of impedance control on a treadmill walking test where the exoskeleton's user interacts with a virtual teacher providing pre-recorded reference trajectories.

\begin{figure*}[ht]
    \centering 
    \includegraphics[width=0.95\linewidth]{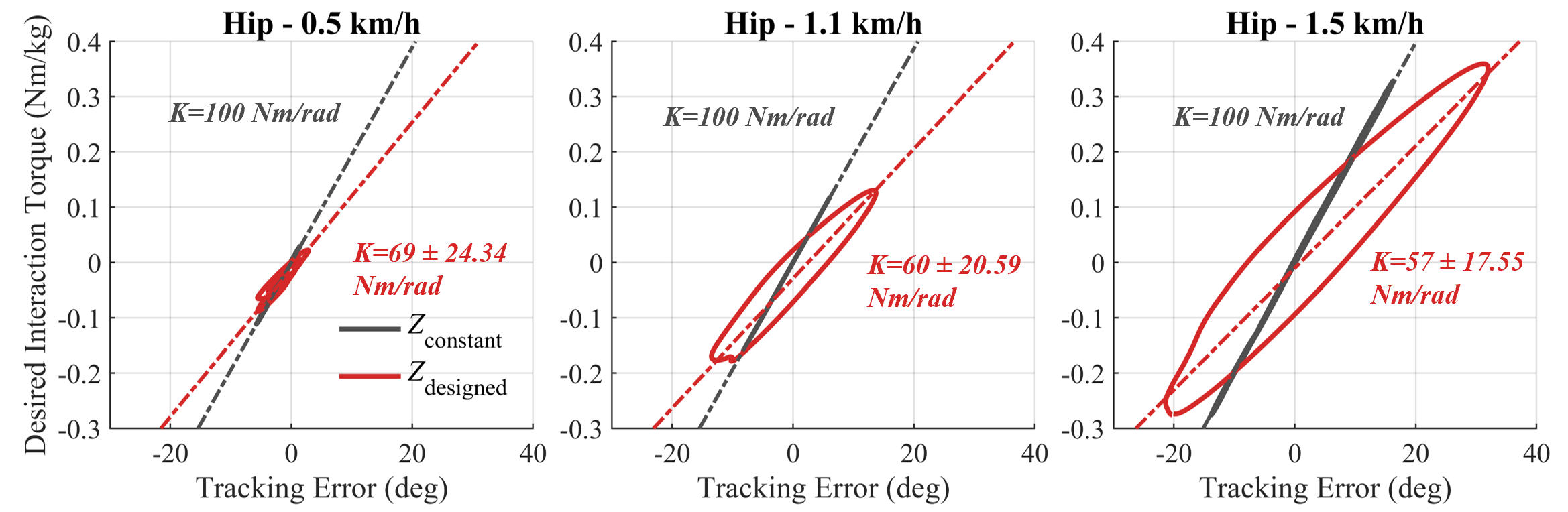}
    \caption{\small Mean impedance variation during the treadmill walking test interacting with a virtual partner at different speeds (from left to right, 0.5 km/h, 1.1 km/h, and 1.5 km/h). On the $y$-axis, the desired interaction torque (impedance controller output) is reported over the tracking error (impedance controller input) on the $x$-axis. For each speed, the $Z_{\text{constant}}$ and the $Z_{\text{designed}}$ 
    are shown together. In dotted line, the average $K$  +/- standard deviation is reported.}
    \label{fig:Effective K}
    \vspace{-0.5cm}
\end{figure*}

Fig.~\ref{fig:Effective K} shows the stiffness at the hip when implementing the impedance laws $Z_{\text{constant}}$ and $Z_{\text{designed}}$ at different walking speeds. 
The slope of the curve (i.e., the mean desired interaction torque $\tau_{\text{des}}$ over the mean tracking error $\theta_{\text{e}}$) is the actual impedance value computed by the impedance controller.
Each subfigure also has a dotted line showing the linear fit to the data for each of the two impedance laws, and the slopes of these fits correspond to the average stiffness at the hip.
As expected, the stiffness for $Z_{\text{constant}}$ is 100~Nm/rad at each of the three walking speeds, while the average stiffness for $Z_{\text{designed}}$ is between 50~Nm/rad and 100~Nm/rad, which represent the commanded stiffnesses at high and low frequencies, respectively. The average stiffness decreases with increased walking speed (69~Nm/rad at 0.5~km/h, 60~Nm/rad at 1.1~km/h, 57~Nm/rad at 1.5~km/h), demonstrating the reduced stiffness at higher frequencies.
Compared to $Z_{\text{constant}}$, the impedance law $Z_{\text{designed}}$ was capable of conveying the same amount of torque (maximum desired interaction torque values), while it reduced the maximum desired torque rate of change (in Nm(kg s)$^{-1}$) by 12\% at 0.5~km/h, 14\% at 1.1 km/h, and 17\% at 1.5 km/h. This is important in the case of weak patients where a considerable amount of torque is required to sustain movement, but abrupt changes in torque can lead to unwanted spastic responses~\cite{li2015new}.

\subsection{IMU-Exoskeleton Human-Robot-Human Interaction}

In this section, we present and discuss the results of the three experiments proposed to evaluate the IMU-exoskeleton unidirectional pHRHI. The three walking tasks (normal walking, asymmetric walking, and walking in the presence of obstacles) were performed on two treadmills where the student and teacher walked without seeing each other.

\begin{table}[t]
    \centering
    \caption{\small Average correlation coefficients between the steps of the teacher and the student during the treadmill walking test.}
    \renewcommand{\arraystretch}{1.3}
    \begin{tabular}{l|c c cc}
        \hline
         & No Int. & Soft Int. & Stiff Int.\\ [0.5ex]
        \hline
        $\overline{\rho}_{\text{hip}}$ & $-$0.00$\pm$0.556 & 0.41$\pm$0.243 & 0.61$\pm$0.158 \\ 
        $\overline{\rho}_{\text{knee}}$ & $-$0.01$\pm$0.362 & 0.25$\pm$0.234 & 0.28$\pm$0.224 \\ 
        $\overline{\rho}_{\text{ankle}}$ & 0.00$\pm$0.3 & 0.35$\pm$0.221 & 0.24$\pm$0.298 \\ [0.3ex]
        \hline
    \end{tabular}
    \label{tab:Correlation IMU table}
    \vspace{-0.45cm}
\end{table}

\begin{figure}[ht]
    \centering 
    \vspace{0.2cm}
    \includegraphics[width=0.95\linewidth]{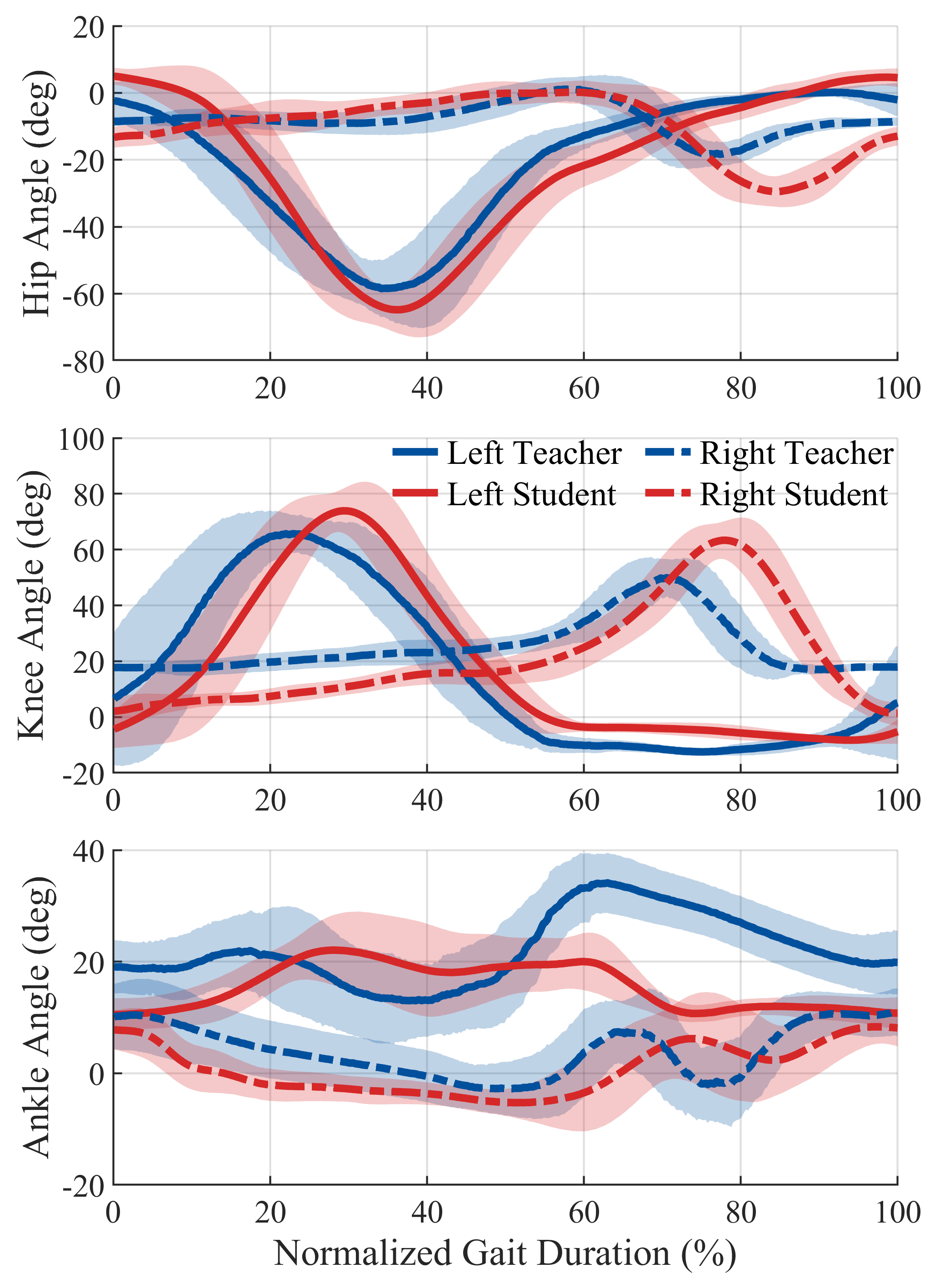}
    \caption{\small Mean +/- standard deviation of the teacher and the student joint angles during the teacher-student haptic information transfer test. The asymmetric gait pattern is rendered through the virtual IMU-exoskeleton interaction.}
    \label{fig:IMU Asymmetric Gait}
    \vspace{-0.4cm}
\end{figure}

\begin{figure}[t]
    \centering 
    \vspace{0.2cm}
    \includegraphics[width=1\linewidth]{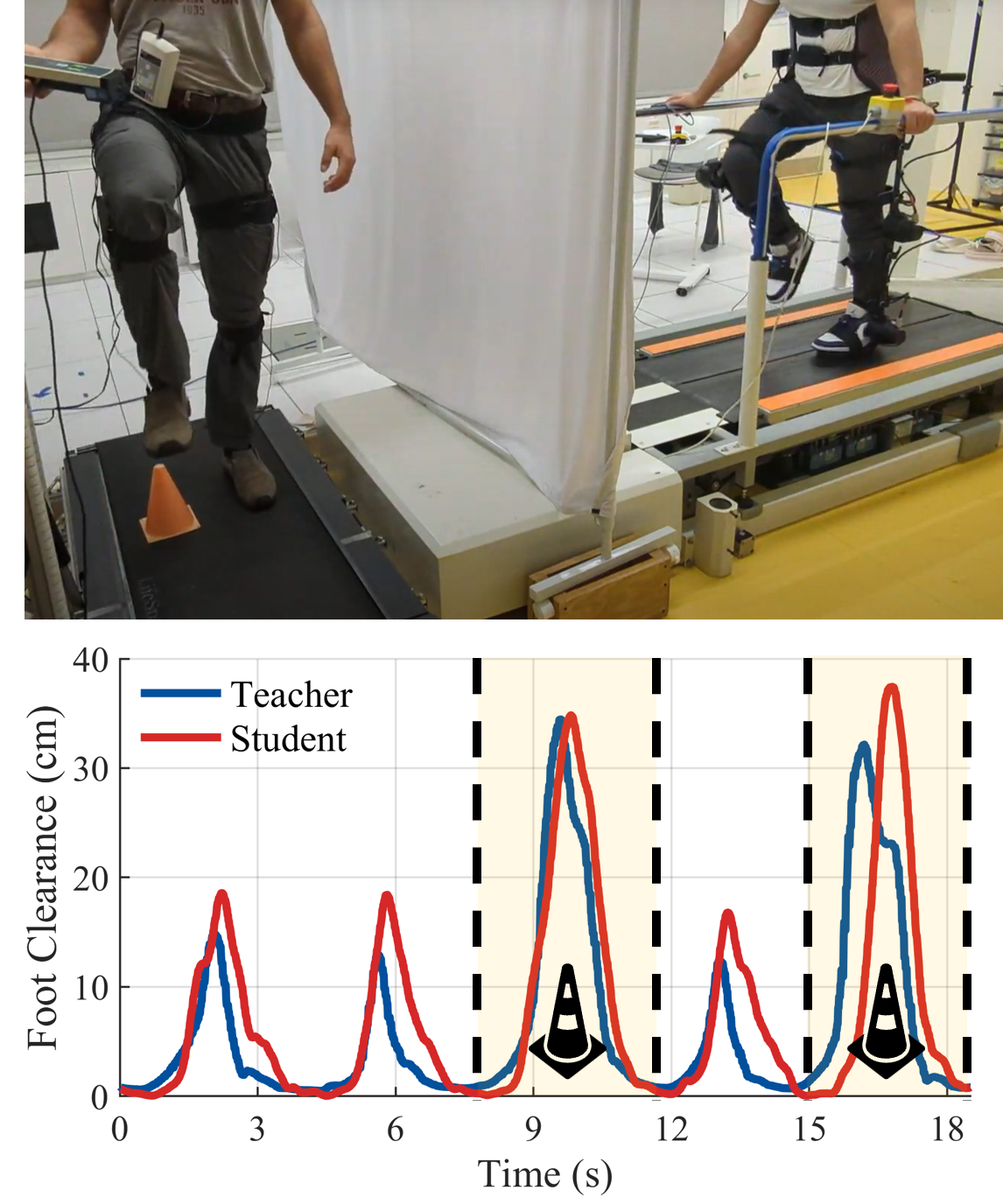}
    \caption{\small On top, a snapshot of the treadmill walking test with random obstacles where the teacher (on the left) avoids an obstacle, while the student (on the right) successfully replicates the teacher's movement only relying on the haptic feedback. On bottom, foot clearance during a section of the test. The obstacles traversed by the teacher (yellow-shaded steps) are perceived by the student.}
    \label{fig:Obstacles}
    \vspace{-0.4cm}
\end{figure}

For the first experiment, the average correlation coefficients between the teacher's and student's joints were evaluated to see if a stiffer connection results in smaller differences between the joint configurations~\cite{short2023haptic} (Table~\ref{tab:Correlation IMU table}). In the no interaction case $Z_{\text{zero}}$, there is no correlation between the two subjects ($\overline{\rho}\approx0$) and large variability due to the lack of synchronization and different frequencies of the gait strides. During the interaction tests with $Z_{\text{soft}}$ and $Z_{\text{stiff}}$, the hip joints showed an increased correlation and a reduction of variability as the connection stiffness increased due to the gait synchronization induced by the virtual connection ($\overline{\rho}_{\text{hip}}$ from 0.41$\pm$0.243 to 0.61$\pm$0.158). We observed similar results for the knee joint, though the magnitude of synchronization was less evident ($\overline{\rho}_{\text{knee}}$ from 0.25$\pm$0.234 to 0.28$\pm$0.224). For the ankle, it is still possible to observe the synchronization of trajectories ($\overline{\rho}_{\text{ankle}}>0$), but not the increasing correlation trend across stiffnesses observed in the other joints.

These results suggest that the interaction control managed to transfer different degrees of haptic guidance about the desired walking pattern from the teacher to the student depending on the strength of the impedance~\cite{short2023haptic}. Specifically, in the no connection case $Z_{\text{zero}}$, there was a complete de-synchronization in the trajectories of the subjects (zero average correlation and high variability, due to the large standard deviation values), as no haptic suggestions were transmitted. Conversely, in both the soft and stiff connection cases, the trajectories of the two subjects exhibited greater synchronization. Moreover, increasing the stiffness of the connection resulted in a larger average correlation and less variability indicated by the smaller standard deviation values. Still, the average correlation values were all under 0.6, meaning that an even stiffer connection is needed to completely transmit the teacher's movements, especially when the student is a patient with very limited residual movement. 

In the second experiment, we evaluated the capability of the controller to transmit physical cues from the teacher to the student about a novel gait pattern, and the student's adaptation to this walking pattern. To do so, we compared the joint kinematics during the teacher-student haptic information transfer test as shown in Fig.~\ref{fig:IMU Asymmetric Gait}, both for the teacher and the student. The teacher performed an asymmetric gait, with differences between his left and right legs during peak flexion around 40 deg, 16 deg, and 23 deg for the hip, knee, and ankle joints, respectively. We observed that the student followed the IMU-reference via the virtual interaction, without prior knowledge of this new gait pattern and without visual feedback of the teacher movements. The teacher rendered the asymmetric gait with differences between his left and right legs during peak flexion around 35 deg, 11 deg, and 14 deg for the hip, knee, and ankle joints, respectively.
The virtual interaction allowed the teacher's suggestions about this modified gait pattern to be transmitted through haptic feedback, resulting in the student performing the same asymmetric gait as the teacher. In a clinical setting, this framework could allow a therapist to apply corrections to multiple joints while teaching how to properly perform dynamic tasks and to switch between different rehabilitation exercises (e.g. stairs, ramps) while treating their patient.

Finally, Fig.~\ref{fig:Obstacles} shows the foot clearance of the right leg during a section of the treadmill walking test with two random obstacles. In this example, we observed that the teacher increased his foot clearance by 21~cm and 20~cm to avoid the two obstacles. The increased foot clearance was matched by the student, as he increased his own clearance to 17~cm and 20~cm respectively, while relying only on the haptic feedback. The virtual interaction between the teacher and the student was capable of promptly communicating the presence of the obstacles from the teacher to the student. The kinematics differences in the two subjects are due to the compliant nature of the interaction, and the fact that only haptic feedback was used to inform the student during this test.
These results suggest that the proposed framework can handle unstructured scenarios (e.g., obstacle avoidance). Moreover, this can be adapted to more complex or unpredictable situations that occur in rehabilitation settings which a conventional approach of using predefined trajectories cannot handle. Additionally, the concept of virtual obstacles enables the promotion of variability in the patient's movement without exposing them to physical obstacles that could compromise safety.

\section{Conclusions and Future Work}
This work presents a new framework for lower-limb rehabilitation where a teacher, wearing an IMU suit, can remotely virtually interact with a student wearing a lower-limb exoskeleton.
To achieve the virtual interaction, we presented an impedance controller to allow the teacher to guide the student. 
The proposed IMU-exoskeleton interaction was found capable of effectively transferring haptic information from teacher to student, enabling parallel multi-joint mobilization both in the classical (i.e., treadmill walking) and more unstructured (e.g., virtual obstacle avoidance) scenarios.
 
Future studies will transfer this framework to a rehabilitation setting, where teacher and student become therapist and patient (e.g., post-stroke and traumatic spinal cord injury patients) to see if this haptic interaction affects task performance and motor learning, and how a therapist can modulate the robot-mediated assistance and adapt to the patient's capabilities to improve the outcomes of rehabilitation compared to other approaches detailed in Table~\ref{tab:rehabilitation comparison}, namely conventional therapy and exoskeleton therapy. This will allow more intuitive rehabilitation scenarios, addressing patient-specific needs and leveraging the physiotherapist's expertise.

\bibliographystyle{IEEEtran}
\bibliography{IEEEabrv,main}

\begin{thebibliography}{10}
\providecommand{\url}[1]{#1}
\csname url@samestyle\endcsname
\providecommand{\newblock}{\relax}
\providecommand{\bibinfo}[2]{#2}
\providecommand{\BIBentrySTDinterwordspacing}{\spaceskip=0pt\relax}
\providecommand{\BIBentryALTinterwordstretchfactor}{4}
\providecommand{\BIBentryALTinterwordspacing}{\spaceskip=\fontdimen2\font plus
\BIBentryALTinterwordstretchfactor\fontdimen3\font minus
  \fontdimen4\font\relax}
\providecommand{\BIBforeignlanguage}[2]{{%
\expandafter\ifx\csname l@#1\endcsname\relax
\typeout{** WARNING: IEEEtran.bst: No hyphenation pattern has been}%
\typeout{** loaded for the language `#1'. Using the pattern for}%
\typeout{** the default language instead.}%
\else
\language=\csname l@#1\endcsname
\fi
#2}}
\providecommand{\BIBdecl}{\relax}
\BIBdecl

\bibitem{belda2011rehabilitation}
J.-M. Belda-Lois, S.~Mena-del Horno, I.~Bermejo-Bosch, J.~C. Moreno, J.~L.
  Pons, D.~Farina, M.~Iosa, M.~Molinari, F.~Tamburella, A.~Ramos \emph{et~al.},
  ``Rehabilitation of gait after stroke: a review towards a top-down
  approach,'' \emph{Journal of NeuroEngineering and Rehabilitation}, vol.~8,
  pp. 1--20, 2011.

\bibitem{mccrory2014work}
B.~McCrory, J.~M. Burnfield, A.~R. Darragh, J.~L. Meza, S.~L. Irons,
  P.~Chernyavskiy, A.~M. Link, and G.~Brusola, ``Work injuries among therapists
  in physical rehabilitation,'' in \emph{Proceedings of the Human Factors and
  Ergonomics Society Annual Meeting}, vol.~58, no.~1.\hskip 1em plus 0.5em
  minus 0.4em\relax SAGE Publications Sage CA: Los Angeles, CA, 2014, pp.
  1072--1076.

\bibitem{moeller2023use}
T.~Moeller, F.~Moehler, J.~Krell-Roesch, M.~De{\v{z}}man, C.~Marquardt,
  T.~Asfour, T.~Stein, and A.~Woll, ``Use of lower limb exoskeletons as an
  assessment tool for human motor performance: a systematic review,''
  \emph{Sensors}, vol.~23, no.~6, p. 3032, 2023.

\bibitem{gassert2018rehabilitation}
R.~Gassert and V.~Dietz, ``Rehabilitation robots for the treatment of
  sensorimotor deficits: a neurophysiological perspective,'' \emph{Journal of
  NeuroEngineering and Rehabilitation}, vol.~15, no.~1, pp. 1--15, 2018.

\bibitem{marchal2019haptic}
L.~Marchal-Crespo, P.~Tsangaridis, D.~Obwegeser, S.~Maggioni, and R.~Riener,
  ``Haptic error modulation outperforms visual error amplification when
  learning a modified gait pattern,'' \emph{Frontiers in Neuroscience},
  vol.~13, p.~61, 2019.

\bibitem{hobbs2020review}
B.~Hobbs and P.~Artemiadis, ``A review of robot-assisted lower-limb stroke
  therapy: unexplored paths and future directions in gait rehabilitation,''
  \emph{Frontiers in Neurorobotics}, vol.~14, p.~19, 2020.

\bibitem{baud2021review}
R.~Baud, A.~R. Manzoori, A.~Ijspeert, and M.~Bouri, ``Review of control
  strategies for lower-limb exoskeletons to assist gait,'' \emph{Journal of
  NeuroEngineering and Rehabilitation}, vol.~18, no.~1, pp. 1--34, 2021.

\bibitem{marchal2009review}
L.~Marchal-Crespo and D.~J. Reinkensmeyer, ``Review of control strategies for
  robotic movement training after neurologic injury,'' \emph{Journal of
  NeuroEngineering and Rehabilitation}, vol.~6, no.~1, pp. 1--15, 2009.

\bibitem{celian2021day}
C.~Celian, V.~Swanson, M.~Shah, C.~Newman, B.~Fowler-King, S.~Gallik,
  K.~Reilly, D.~J. Reinkensmeyer, J.~Patton, and M.~R. Rafferty, ``A day in the
  life: a qualitative study of clinical decision-making and uptake of
  neurorehabilitation technology,'' \emph{Journal of NeuroEngineering and
  Rehabilitation}, vol.~18, no.~1, pp. 1--12, 2021.

\bibitem{hasson2023neurorehabilitation}
C.~J. Hasson, J.~Manczurowsky, E.~C. Collins, and M.~Yarossi,
  ``Neurorehabilitation robotics: how much control should therapists have?''
  \emph{Frontiers in Human Neuroscience}, vol.~17, p. 1179418, 2023.

\bibitem{Kucuktabak2021}
E.~B. K{\"u}{\c{c}}{\"u}ktabak, S.~J. Kim, Y.~Wen, K.~Lynch, and J.~L. Pons,
  ``Human-machine-human interaction in motor control and rehabilitation: a
  review,'' \emph{Journal of NeuroEngineering and Rehabilitation}, vol.~18,
  no.~1, p. 183, Dec 2021.

\bibitem{ganesh2014two}
G.~Ganesh, A.~Takagi, R.~Osu, T.~Yoshioka, M.~Kawato, and E.~Burdet, ``Two is
  better than one: Physical interactions improve motor performance in humans,''
  \emph{Scientific Reports}, vol.~4, no.~1, pp. 1--7, 2014.

\bibitem{short2023haptic}
M.~R. Short, D.~Ludvig, E.~B. K{\"u}{\c{c}}{\"u}ktabak, Y.~Wen, L.~Vianello,
  E.~J. Perreault, L.~Hargrove, K.~Lynch, and J.~L. Pons, ``Haptic human-human
  interaction during an ankle tracking task: Effects of virtual connection
  stiffness,'' \emph{IEEE Transactions on Neural Systems and Rehabilitation
  Engineering}, 2023.

\bibitem{kuccuktabak2023virtual}
E.~B. K{\"u}{\c{c}}{\"u}ktabak, Y.~Wen, M.~Short, E.~Demirba{\c{s}}, K.~Lynch,
  and J.~Pons, ``Virtual physical coupling of two lower-limb exoskeletons,'' in
  \emph{2023 International Conference on Rehabilitation Robotics
  (ICORR)}.\hskip 1em plus 0.5em minus 0.4em\relax IEEE, 2023, pp. 1--6.

\bibitem{rahman2011tele}
M.~H. Rahman, T.~K-Ouimet, M.~Saad, J.~P. Kenn{\'e}, and P.~S. Archambault,
  ``Tele-operation of a robotic exoskeleton for rehabilitation and passive arm
  movement assistance,'' in \emph{2011 IEEE International Conference on
  Robotics and Biomimetics}.\hskip 1em plus 0.5em minus 0.4em\relax IEEE, 2011,
  pp. 443--448.

\bibitem{sakurai2009development}
T.~Sakurai and Y.~Sankai, ``Development of motion instruction system with
  interactive robot suit hal,'' in \emph{2009 IEEE International Conference on
  Robotics and Biomimetics (ROBIO)}.\hskip 1em plus 0.5em minus 0.4em\relax
  IEEE, 2009, pp. 1141--1147.

\bibitem{vianello2024exoskeleton}
L.~Vianello, E.~B. K{\"u}{\c{c}}{\"u}ktabak, M.~Short, C.~Lhoste, L.~Amato,
  K.~Lynch, and J.~Pons, ``Exoskeleton-mediated physical human-human
  interaction for a sit-to-stand rehabilitation task,'' in \emph{2024 IEEE
  International Conference on Robotics and Automation (ICRA)}.\hskip 1em plus
  0.5em minus 0.4em\relax IEEE, 2024, pp. 4521--4527.

\bibitem{pavon2020iot}
N.~Pav{\'o}n-Pulido, J.~A. L{\'o}pez-Riquelme, and J.~J. Feli{\'u}-Batlle,
  ``Io{T} architecture for smart control of an exoskeleton robot in
  rehabilitation by using a natural user interface based on gestures,''
  \emph{Journal of Medical Systems}, vol.~44, pp. 1--10, 2020.

\bibitem{duan2018motion}
P.~Duan, Z.~Duan, S.~Li, and Y.~Chen, ``Motion prediction and delay
  compensation for improved teleoperation of an exoskeletal robot,''
  \emph{Advances in Mechanical Engineering}, vol.~10, no.~2, p.
  1687814018760353, 2018.

\bibitem{koh2021exploiting}
M.~H. Koh, S.-C. Yen, L.~Y. Leung, S.~Gans, K.~Sullivan, Y.~Adibnia, M.~Pavel,
  and C.~J. Hasson, ``Exploiting telerobotics for sensorimotor rehabilitation:
  a locomotor embodiment,'' \emph{Journal of NeuroEngineering and
  Rehabilitation}, vol.~18, pp. 1--21, 2021.

\bibitem{tran2016evaluation}
H.~T. Tran, H.~Cheng, H.~Rui, X.~Lin, M.~K. Duong, and Q.~Chen, ``Evaluation of
  a fuzzy-based impedance control strategy on a powered lower exoskeleton,''
  \emph{International Journal of Social Robotics}, vol.~8, pp. 103--123, 2016.

\bibitem{huo2021impedance}
W.~Huo, H.~Moon, M.~A. Alouane, V.~Bonnet, J.~Huang, Y.~Amirat,
  R.~Vaidyanathan, and S.~Mohammed, ``Impedance modulation control of a
  lower-limb exoskeleton to assist sit-to-stand movements,'' \emph{IEEE
  Transactions on Robotics}, vol.~38, no.~2, pp. 1230--1249, 2021.

\bibitem{baris2024haptic}
E.~B. Küçüktabak, Y.~Wen, S.~J. Kim, M.~R. Short, D.~Ludvig, L.~Hargrove,
  E.~J. Perreault, K.~M. Lynch, J.~L. Pons \emph{et~al.}, ``Haptic transparency
  and interaction force control for a lower-limb exoskeleton,'' \emph{IEEE
  Transactions on Robotics}, 2024.

\bibitem{louie2015gait}
D.~R. Louie, J.~J. Eng, T.~Lam, and S.~C. I. R. E. S.~R. Team, ``Gait speed
  using powered robotic exoskeletons after spinal cord injury: a systematic
  review and correlational study,'' \emph{Journal of NeuroEngineering and
  Rehabilitation}, vol.~12, pp. 1--10, 2015.

\bibitem{perry1995classification}
J.~Perry, M.~Garrett, J.~K. Gronley, and S.~J. Mulroy, ``Classification of
  walking handicap in the stroke population,'' \emph{Stroke}, vol.~26, no.~6,
  pp. 982--989, 1995.

\bibitem{mentiplay2018lower}
B.~F. Mentiplay, M.~Banky, R.~A. Clark, M.~B. Kahn, and G.~Williams, ``Lower
  limb angular velocity during walking at various speeds,'' \emph{Gait \&
  Posture}, vol.~65, pp. 190--196, 2018.

\bibitem{li2015new}
S.~Li and G.~E. Francisco, ``New insights into the pathophysiology of
  post-stroke spasticity,'' \emph{Frontiers in human neuroscience}, vol.~9, p.
  192, 2015.

\end{thebibliography}

\end{document}